\documentclass{article}

\usepackage[colorlinks]{hyperref}       % hyperlinks
\usepackage{amsfonts}       % blackboard math symbols
\usepackage{graphicx}
\usepackage{listings}
\usepackage{multirow}
\usepackage{multicol}
\usepackage{subcaption}
\usepackage{arxiv}
\usepackage{euscript}
\usepackage{amsmath} 
\usepackage[utf8]{vietnam}
\usepackage[english]{babel}

\begin{document}
\title{Towards Comprehensive Vietnamese Retrieval-Augmented Generation and Large Language Models}
%
% \titlerunning{An Efficient  Transformer using Bidirectional Feature Pyramid Network}
\author{{Nguyen Quang Duc} \\
Foundation Models Lab, BKAI\\
Hanoi University of Science and Technology\\
\texttt{ducnq.204876@sis.hust.edu.vn} \\
\And
{Le Hai Son} \\
Foundation Models Lab, BKAI\\
Hanoi University of Science and Technology\\
\texttt{haison.le001@gmail.com} \\
\And
{Nguyen Duc Nhan} \\
University of Information Technology\\
Vietnam National University HCMC\\
\texttt{21520373@gm.uit.edu.vn} \\
\And
{Nguyen Dich Nhat Minh} \\
Foundation Models Lab, BKAI\\
Hanoi University of Science and Technology\\
\texttt{minh.ndn215429@sis.hust.edu.vn} \\
\And
{Le Thanh Huong} \\
Foundation Models Lab, BKAI\\
Hanoi University of Science and Technology\\
\texttt{huonglt@soict.hust.edu.vn} \\
\And
{Dinh Viet Sang} \thanks{Corresponding author}\\
Foundation Models Lab, BKAI\\
Hanoi University of Science and Technology\\
\texttt{sangdv@soict.hust.edu.vn} \\
%% \AND
%% Coauthor \\
%% Affiliation \\
%% Address \\
%% \texttt{email} \\
%% \And
%% Coauthor \\
%% Affiliation \\
%% Address \\
%% \texttt{email} \\
%% \And
%% Coauthor \\
%% Affiliation \\
%% Address \\
%% \texttt{email} \\
}

% Uncomment to override  the `A preprint' in the header
%\renewcommand{\headeright}{Technical Report}
%\renewcommand{\undertitle}{Technical Report}
\renewcommand{\shorttitle}{Towards Comprehensive Vietnamese RAG and LLMs}

\maketitle              % typeset the header of the contribution
\begin{abstract}
This paper presents our contributions towards advancing the state of Vietnamese language understanding and generation through the development and dissemination of open datasets and pre-trained models for Vietnamese Retrieval-Augmented Generation (RAG) and Large Language Models (LLMs).
\keywords{RAG, LLMs, Open Datasets}
\end{abstract}
\section{Introduction}
We hope that the research community, both in Vietnam and around the world, will join forces in the endeavor to construct large and high-quality datasets for the training and evaluation of Retrieval-Augmented Generation (RAG) and Large Language Models (LLMs) for Vietnamese. By collaborating on this front, we can collectively push the boundaries of what's possible in natural language processing for Vietnamese, unlocking new opportunities for innovation and application in the field.

Together, let's work towards an open scientific community that benefits everyone!
\section{Main contributions}
Our main contributions are as follows:
\begin{itemize}
    \item A massive Vietnamese NewsCorpus dataset of around 32M articles, a substantial 53 GB in size, rigorously cleaned, deduplicated, and formatted specifically for the continual pretraining of LLMs.
    \item An extensive Vietnamese NewsSapo dataset, structured in a ``title-abstract-contents'' format, specifically designed to enhance the training of sentence/passage embeddings.
    \item An additional large-scale Vietnamese NewsCategory dataset in a ``text-category'' format, specifically designed for text classification tasks.
    \item Vietnamese Alpaca datasets, tailored for supervised fine-tuning LLMs.
    \item Synthetic self-chat and roleplay realm datasets are developed for enhancing the conversation capability of LLMs through supervised fine-tuning.
    \item A good Vietnamese bi-encoder model is presented for advanced sentence embedding tasks.
    \item We also offer two base models, Vietnamese LLaMA2-7b, which have been further pretrained on an expansive corpus of Vietnamese text, 40 GB and 120 GB, respectively, derived from LLaMA2, marking a significant advancement in the understanding and generation of the Vietnamese language.
\end{itemize}

\section{Details}

\subsection{Vietnamese NewsCorpus dataset}
\label{newcoprus}
The Binhvq News Corpus \cite{binhvq}, a widely used dataset featuring approximately 20 million articles from diverse sources, received its last update in May 2021. To enhance this collection, we gathered an additional 10 million articles up until November 2023. By integrating these newly acquired articles with the existing Binhvq News Corpus, we have created an extensive Vietnamese News Corpus comprising about 32M articles. Subsequent fuzzy deduplication was conducted to remove duplicate articles, resulting in 53 GB of clean data, which is ready for the continual pretraining of LLMs.
    
\textbf{Huggingface link:} \url{https://huggingface.co/datasets/bkai-foundation-models/BKAINewsCorpus}

\subsection{Vietnamese NewsSapo dataset}
The Vietnamese NewsSapo dataset was constructed to train sentence/passage embeddings. Our dataset is structured in a "title-abstract-contents" format, where each news article is represented by a tuple of (title, abstract, content). The content is the main text body of the article and has been processed to remove images, videos, and other non-textual elements. The dataset contains 31,728,183 triples.

\textbf{Huggingface link:} \url{https://huggingface.co/datasets/bkai-foundation-models/NewsSapo}
    
\subsection{Vietnamese NewsCategory dataset}
% The Vietnamese NewsCategory dataset is collected from the Vnexpress news website and is extracted for clustering tasks, which is similar to the THUCTC dataset \cite{thuctc}.

The Vietnamese NewsCategory dataset is constructed similarly to the THUCTC dataset \cite{thuctc}. The dataset is collected from VnExpress and is extracted for classification tasks. It contains 596,524 samples, each of which consists of five fields: id (index), title, sapo (summary), content (the contents of articles), and label (article topic). 

The articles are categorized into 21 topics, including: Celebrities (Ngôi Sao), World (Thế giới), Youth Entertainment (Giải trí giới trẻ), Sports (Thể thao), Business (Kinh doanh), Health (Sức khỏe), Current Affairs (Thời sự), Entertainment (Giải trí), Confession (Tâm sự), Legal (Pháp luật), Science (Khoa học), Digital (Số hóa), Education (Giáo dục), Travel (Du lịch), Cars (Xe), Life (Đời sống), Relaxation (Thư giãn), Real Estate (Bất động sản), Opinion (Ý kiến), Podcasts, and Perspective (Góc nhìn).

% Note: 'Ngôi sao' articles are written in a separate section, while 'Giải trí giới trẻ' articles are rewritten from 'iOne' section.

\textbf{Huggingface link:} \url{https://huggingface.co/datasets/bkai-foundation-models/NewsCategory}

\subsection{Vietnamese Alpaca datasets}
% Sơn
\subsubsection{Standard Vietnamese Alpaca}
This dataset is specifically tailored for Vietnamese, drawing inspiration from the methodologies of Stanford Alpaca \cite{taori2023alpaca} and Self-Instruct \cite{wang2022self}.
The construction of this dataset involved a systematic two-step process:

\begin{itemize}
    \item \textbf{Creation of Vietnamese Seed Tasks:} Following the idea in Self-Instruct \cite{wang2022self}, we carefully developed a wide-ranging collection of seed tasks for Vietnamese. This was achieved using GPT-4, alongside meticulous manual crafting.
    \item \textbf{Generation of Instructions:} Leveraging the seed tasks prepared in the first step, we engaged in the instruction generation technique inspired by Stanford Alpaca \cite{taori2023alpaca}. Utilizing GPT-4, GPT-3.5 turbo, and GPT-3.5-instruct, we generated 50K instructions. This process was carried out using various configurations to ensure the production of a rich and diverse array of linguistic scenarios.
\end{itemize}

% Lastly, we also build a scoring system to do crowdsourced evaluation and filter/re-generate for better quality.

\textbf{Huggingface link:} \url{https://huggingface.co/datasets/bkai-foundation-models/vi-alpaca}

\subsubsection{Modified Vietnamese Alpaca}
This dataset has been developed using a similar method to that of the Standard Vietnamese Alpaca, with a notable difference in format. While the Standard Vietnamese Alpaca adopts the ``instructions/inputs/outputs'' format as suggested by \cite{taori2023alpaca}, our method, influenced by \cite{cui2023efficient}, utilizes the ``input/output'' format. This alteration allows for the generation of more diverse samples and more lengthy output, employing GPT-4, GPT-3.5, and GPT-3.5-instruct. The resulting Modified Vietnamese Alpaca dataset contains 25K samples.

\textbf{Huggingface link:} \url{https://huggingface.co/datasets/bkai-foundation-models/vi-alpaca-input-output-format}

\subsection{Vietnamese Self-chat dataset}
% Sơn
This dataset contains around 30K dialogues designed to enhance the model's ability to engage in multi-turn conversations with humans. We follow two steps:

\begin{itemize}
    \item \textbf{Instruction Generation:} We employ the methodology outlined in Self-Instruct \cite{wang2022self} to craft a diverse set of instructions.
    \item \textbf{Synthetic Self-Chat Conversations:} Building upon the instructions generated in the first step, we draw inspiration from Baize \cite{xu2023baize} to generate synthetic multi-turn dialogues. These simulations serve as practical scenarios for the model to learn from and adapt to dynamic conversation flows.
\end{itemize}

% \textbf{Step 1: }

%  %This paper serves as a guide for aligning pretrained language models with specific instructions, providing a structured foundation for subsequent dialogue generation.

% \textbf{Step 2: }

By combining these two steps, we aim to create a robust and versatile dataset that empowers the model to navigate and contribute effectively in complex conversational scenarios. This dataset serves as a valuable resource for refining the model's language understanding and response generation capabilities in the context of human-like dialogue.

\textbf{Huggingface link:} \url{https://huggingface.co/datasets/bkai-foundation-models/vi-self-chat-sharegpt-format}

\subsection{Vietnamese Roleplay Realm dataset}
% Minh
This is a dataset of GPT-generated characters made to increase the ability of open-source language models to roleplay. It contains 446 characters generated by GPT-3.5. The total number of dialogues is about 9K.

To construct this dataset, we follow four steps:

\begin{itemize}
    \item \textbf{Character Generation:} Creates a set of fictional characters with GPT-3.5 based on a prompt and a seed list of characters. The generated output fields for each character are ``name'', ``context'', ``greeting'', and ``example\_dialogue''.
    \item \textbf{Topic Generation:} We then created conversation topics for each character, drawing from their descriptions. The output field for this step is ``topics''. We generate 20 topics for each character.
    \item \textbf{Dialogue generation: } We generated dialogues Based on the character descriptions and topics. The output for this step is encapsulated in the ``dialogues'' field.
    \item \textbf{Checking and Refining:} Given that the dataset may contain errors in Vietnamese, a review and correction process is necessary to ensure accuracy and refinement.
\end{itemize}

\textbf{Huggingface link:} \url{https://huggingface.co/datasets/bkai-foundation-models/vietnamese-roleplay-realm}

\subsection{Vietnamese Bi-encoder}
% Sang
This is a \href{https://www.SBERT.net}{sentence-transformers} model: It maps sentences \& paragraphs to a 768 dimensional dense vector space and can be used for tasks like clustering or semantic search.

We train the model on a merged training dataset that consists of: 
\begin{itemize}
  \item MS Macro (translated into Vietnamese)
  \item SQuAD v2  (translated into Vietnamese)
  \item 80\% of the training set from the Legal Text Retrieval Zalo 2021 challenge
\end{itemize}

We use \href{https://github.com/VinAIResearch/PhoBERT}{phobert-base-v2} \cite{phobert} as the pre-trained backbone.

Here are the results on the remaining 20\% of the training set from the Legal Text Retrieval Zalo 2021 challenge:

\begin{table}[!ht]
\caption{Results on the Legal Text Retrieval Zalo 2021 challenge}
\centering
\begin{tabular}{l|l|c|c|c|c|c}
\hline
\textbf{Pretrained Model} & \textbf{Training Datasets} & \textbf{Acc@1} & \textbf{Acc@10} & \textbf{Acc@100} & \textbf{Pre@10} & \textbf{MRR@10} \\ \hline
\href{https://huggingface.co/keepitreal/vietnamese-sbert}{Vietnamese-SBERT} & - & 32.34 & 52.97 & 89.84 & 7.05 & 45.30 \\ 
PhoBERT-base-v2 & MSMACRO & 47.81 & 77.19 & 92.34 & 7.72 & 58.37 \\ 
PhoBERT-base-v2 & MSMACRO + SQuADv2.0 + 80\% Zalo & 73.28 & 93.59 & 98.85 & 9.36 & 80.73 \\ \hline
\end{tabular}

\label{table:results}
\end{table}

\textbf{Huggingface link:} \url{https://huggingface.co/bkai-foundation-models/vietnamese-bi-encoder}

\subsection{Vietnamese LLaMA2}
% Sang
\subsubsection{Vietnamese LLaMA2-7b-40Gb}
We employed \href{https://github.com/google/sentencepiece}{SentencePiece} to retrain a Vietnamese tokenizer with a vocabulary size of 20K. No Vietnamese word segmentation was used. We then merged this vocabulary with the original one of LLaMA2, removing duplicate tokens. The new tokenizer significantly improves when encoding Vietnamese text, reducing the number of tokens by 50\% compared to ChatGPT and approximately 70\% compared to the original LLaMA2.

We conducted a single-epoch continual pretraining, also known as incremental pretraining, using the LLaMA2-chat 7B model on a mixed dataset totaling 40.5 GB, comprised of:
\begin{itemize}
    \item 19 GB \href{https://github.com/binhvq/news-corpus}{NewsCorpus} \cite{binhvq}
    \item 1.1 GB Vietnamese Wikipedia
    \item 1.6 GB \href{https://www.kaggle.com/datasets/iambestfeeder/10000-vietnamese-books}{Vietnamese books}
    \item 4.5 GB Vietnamese legal documents (crawled from thuvienphapluat and processed by ourselves)
    \item 2.1 GB Vietnamese legal text (from \href{https://huggingface.co/datasets/c4}{C4-vi})
    \item 1.1 GB English Books (sub-sampled from \href{https://huggingface.co/datasets/pg19}{pg19})
    \item 1.1 GB English Wikipedia (sub-sampled from 20220301.en wikipedia)
    \item 10 GB English Text (sub-sampled from \href{https://huggingface.co/datasets/c4}{C4-en})
\end{itemize}

We trained the model on a DGX A100 system, utilizing four GPU A100 in 10 days (about 1000 GPU hours). 

% Hyperparameters are set as follows:
% \begin{itemize}
%     \item Training Regime: BFloat16 mixed precision
%     \item Lora Config: 
% \end{itemize}

% \begin{verbatim}
% {
%   "base_model_name_or_path": "meta-llama/Llama-2-7b-chat-hf",
%   "bias": "none",
%   "enable_lora": null,
%   "fan_in_fan_out": false,
%   "inference_mode": true,
%   "lora_alpha": 32.0,
%   "lora_dropout": 0.05,
%   "merge_weights": false,
%   "modules_to_save": [
%     "embed_tokens",
%     "lm_head"
%   ],
%   "peft_type": "LORA",
%   "r": 8,
%   "target_modules": [
%     "q_proj",
%     "v_proj",
%     "k_proj",
%     "o_proj",
%     "gate_proj",
%     "down_proj",
%     "up_proj"
%   ],
%   "task_type": "CAUSAL_LM"
% }
% \end{verbatim}

We also provide the \href{https://huggingface.co/bkai-foundation-models/vietnamese-LLaMA2-7b-40GB/tree/main/pt_lora_model}{LoRA part} so that you can integrate it with the original LLaMA2-chat-7b by yourself.

\textbf{Huggingface link:} \url{https://huggingface.co/bkai-foundation-models/vietnamese-llama2-7b-40GB}

\subsubsection{Vietnamese LLaMA2-7b-120GB}
\subsection*{Tokenizer}
We enhance our previous tokenizer in \href{https://huggingface.co/bkai-foundation-models/vietnamese-LLaMA2-7b-40GB}{Vietnamese-LLaMA2-7b-40GB} by training \href{https://github.com/google/sentencepiece}{SentencePiece} on a more extensive collection of clean Vietnamese documents spanning diverse domains such as news, books, stock, finance, and laws. In contrast to the previous version, we follow the original LLaMA-2 paper to split all numbers into individual digits. Again, the updated tokenizer markedly enhances the encoding of Vietnamese text, cutting down the number of tokens by 50\% compared to ChatGPT and approximately 70\% compared to the original LLaMA2.

\subsection*{Pretraining data}
Here are our data sources:
\begin{itemize}
\item Vietnamese NewsCorpus described in Section \ref{newcoprus}
\item 1.3 GB Vietnamese Wikipedia (updated to October 2023)
\item 8.5 GB \href{https://www.kaggle.com/datasets/iambestfeeder/10000-vietnamese-books}{Vietnamese books}
\item 4.8 GB Vietnamese legal documents (clean and dedup)
\item 1.6 GB stock news (clean and dedup)
\item 43 GB Vietnamese text (subsampled from \href{https://huggingface.co/papers/2309.09400}{Culturax-vi \cite{nguyen2023culturax}})
\item 2.3 GB English Books (sub-sampled from \href{https://huggingface.co/datasets/pg19}{pg19})
\item 2.2 GB English Wikipedia
\item 16 GB English text (subsampled from \href{https://huggingface.co/papers/2309.09400}{Culturax-en \cite{nguyen2023culturax}})
\end{itemize}
We then merge all data sources and perform the last deduplication, resulting in a final pretraining dataset of 124 GB, including 104 GB of Vietnamese text and 20 GB of English text. 

\subsection*{Continual pretraining}
We conduct a single-epoch continual pretraining using the LLaMA2-7B model. We trained the model on a DGX A100 system, utilizing four GPU A100 in 40 days (about 4000 GPU hours). 

Hyperparameters are set as the same as the \href{https://huggingface.co/bkai-foundation-models/vietnamese-LLaMA2-7b-40GB}{Vietnamese-LLaMA2-7b-40GB}.

We also provide the \href{https://huggingface.co/bkai-foundation-models/vietnamese-LLaMA2-7b-120GB/tree/main/pt_lora_model}{LoRA part} so that you can integrate it with the original LLaMA2-7b by yourself.

\subsection*{Training loss}
The red line indicates the learning curve of \href{https://huggingface.co/bkai-foundation-models/vietnamese-LLaMA2-7b-40GB}{Vietnamese-LLaMA2-7b-40GB}, while the cyan one corresponds to the new model of 120 GB.
\begin{figure}[!ht]
\centering
\includegraphics[width=0.8\textwidth]{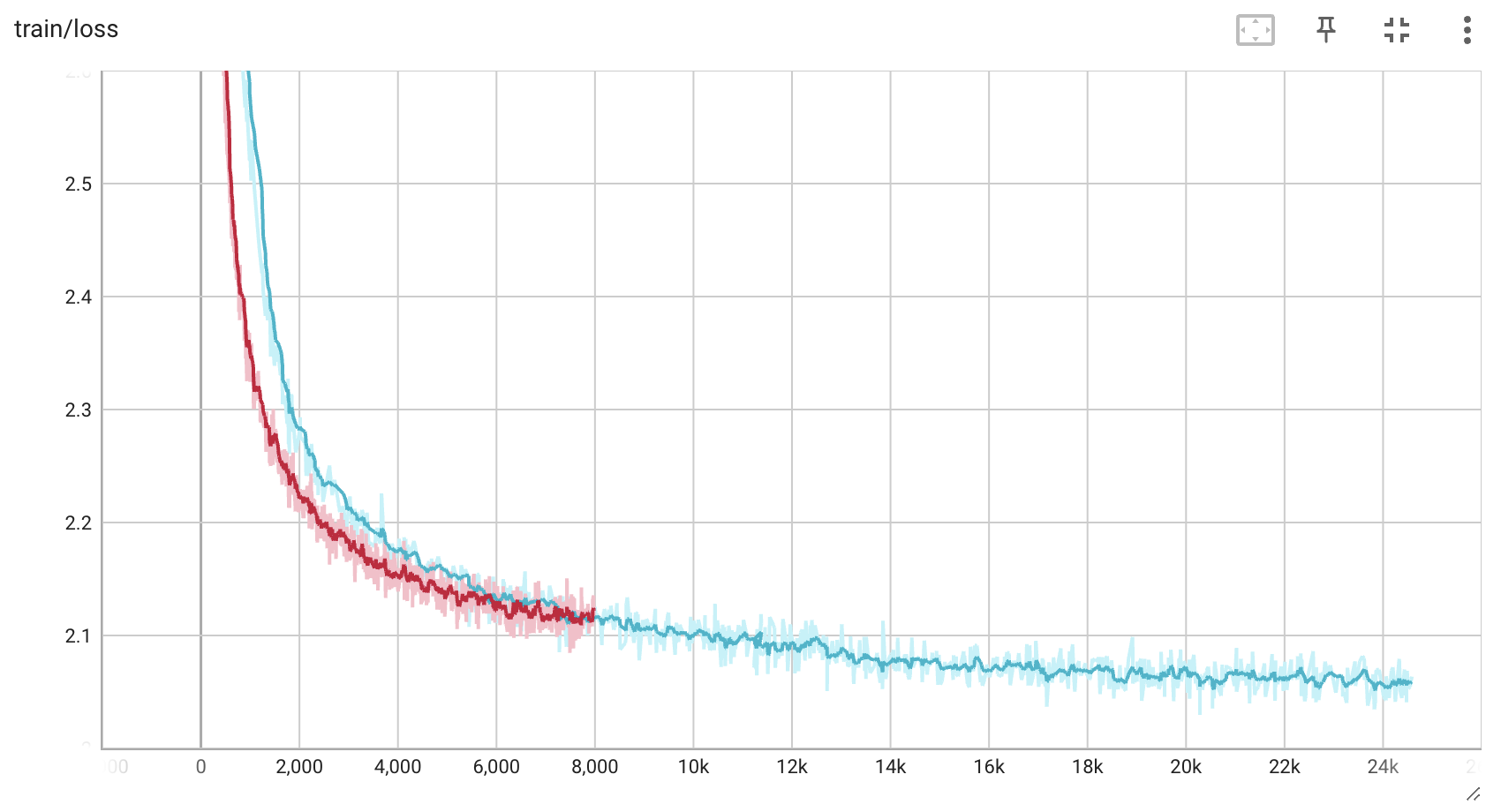}
\end{figure}

\textbf{Huggingface link:} \url{https://huggingface.co/bkai-foundation-models/vietnamese-llama2-7b-120GB}

\section{Conclusion}
By opening access to our datasets and models, we extend an invitation to the broader research community to collaborate with us on the path to developing more inclusive, efficient, and accessible Vietnamese Retrieval-Augmented Generation (RAG) and Large Language Models (LLMs).

Together, we can drive innovation, enhance linguistic inclusivity, and foster a rich ecosystem of NLP tools and technologies that bring substantial benefits to Vietnam.

\label{sec:conclude}

\section{Acknowledgments}
This work was funded by NAVER Corporation. We extend our gratitude to PHPC - Phenikaa University and NVIDIA for their generous provision of computing resources for model training.

\bibliographystyle{unsrt}
\bibliography{arxiv}

\end{document}